\documentclass[11pt]{article}

\usepackage[preprint]{acl}

\usepackage{times}
\usepackage{latexsym}

\usepackage[T1]{fontenc}
\usepackage[utf8]{inputenc}

\usepackage{microtype}

\usepackage{inconsolata}

\usepackage{graphicx}
\usepackage{booktabs}
\usepackage{longtable}
\usepackage{xcolor}
\usepackage{array}
\usepackage{multirow}
\usepackage{amsmath}
\usepackage{hyperref}
\usepackage{longtable}
\usepackage{enumitem}
\usepackage{seqsplit}
\usepackage[most]{tcolorbox}
\usepackage{lipsum}
\usepackage{geometry}
\usepackage{setspace}
\usepackage{tabularx}
\usepackage{algorithm}
\usepackage{algorithmic}
\usepackage{amssymb}
\usepackage{CJKutf8}

\title{JointCQ: Improving Factual Hallucination Detection with Joint Claim and Query Generation}

\author {
    Fan Xu\textsuperscript{1},
    Huixuan Zhang\textsuperscript{1},
    Zhenliang Zhang\textsuperscript{1},
    Jiahao Wang\textsuperscript{2},
    Xiaojun Wan\textsuperscript{1}
\\
    \textsuperscript{1}Wangxuan Institute of Computer Technology, Peking University\\
    \textsuperscript{2}Trustworthy Technology and Engineering Laboratory, Huawei
\\
\small{\texttt{\{xufan2000,wanxiaojun\}@pku.edu.cn,}
\texttt{\{zhanghuixuan,zhenliang\}@stu.pku.edu.cn,}
\texttt{wangjiahao50@huawei.com}}
}

\begin{document}
\maketitle
\begin{abstract}
Current large language models (LLMs) often suffer from hallucination issues, i,e, generating content that appears factual but is actually unreliable. A typical hallucination detection pipeline involves response decomposition (i.e., claim extraction), query generation, evidence collection (i.e., search or retrieval), and claim verification. However, existing methods exhibit limitations in the first two stages, such as context loss during claim extraction and low specificity in query generation, resulting in degraded performance across the hallucination detection pipeline. In this work, we introduce JointCQ\footnote{\url{https://github.com/pku0xff/JointCQ}}, a joint claim-and-query generation framework designed to construct an effective and efficient claim-query generator. Our framework leverages elaborately designed evaluation criteria to filter synthesized training data, and finetunes a language model for joint claim extraction and query generation, providing reliable and informative inputs for downstream search and verification. Experimental results demonstrate that our method outperforms previous methods on multiple open-domain QA hallucination detection benchmarks, advancing the goal of more trustworthy and transparent language model systems.
\end{abstract}

\section{Introduction}
Large Language Models (LLMs) have achieved remarkable success across a wide range of natural language generation (NLG) tasks, including open-domain question answering (QA)~\cite{kamalloo-etal-2023-evaluating}. However, despite their impressive capabilities, LLMs are susceptible to factual hallucinations, where models generate responses that appear plausible but are factually incorrect, as mentioned in multiple previous works~\cite{huang2023survey, ji2023survey, zhang2023siren}. This issue poses significant challenges for users who rely on LLMs for accurate information, raising critical concerns about the reliability and accountability of AI-generated content. As LLMs continue to advance and become increasingly integrated into real-world applications, addressing hallucinations is crucial to ensuring their trustworthiness and practical utility~\cite{pal-etal-2023-med,dahl2024large}. Detecting factual hallucinations in generated content has thus become a critical area of research.

Prior studies have explored various detection methods with distinct limitations. Some approaches rely on self-verification techniques, such as prompting LLMs or sampling generations~\cite{manakul-etal-2023-selfcheckgpt, ni-etal-2024-afacta}, which may inherit the same biases or knowledge gaps as the original model. Others analyze internal model states or generation probabilities~\cite{zhang-etal-2023-enhancing-uncertainty, azaria-mitchell-2023-internal}, but these signals can be opaque and model-specific. In contrast, retrieval-based methods, which systematically search for relevant external information and compare it with generated content, have proven particularly effective, as they provide concrete, verifiable evidence for hallucination detection~\cite{cheng-etal-2024-small, chern2023factool}.
In fields where reliable information is essential, such as healthcare, finance, scientific research, or any scenario involving internal or sensitive data, retrieval-based methods become particularly essential.
Existing retrieval-based detection methods for open-domain question answering typically decompose responses, generate queries and perform evidence retrieval and claim verification. However, these approaches frequently struggle with suboptimal decomposition~\cite{metropolitansky-larson-2025-towards,wanner-etal-2024-closer,ullrich2025claim} and query generation~\cite{jeong-etal-2024-adaptive}, limiting their effectiveness.

\begin{figure*}[ht]
\centering
\includegraphics[width=1.0\textwidth]{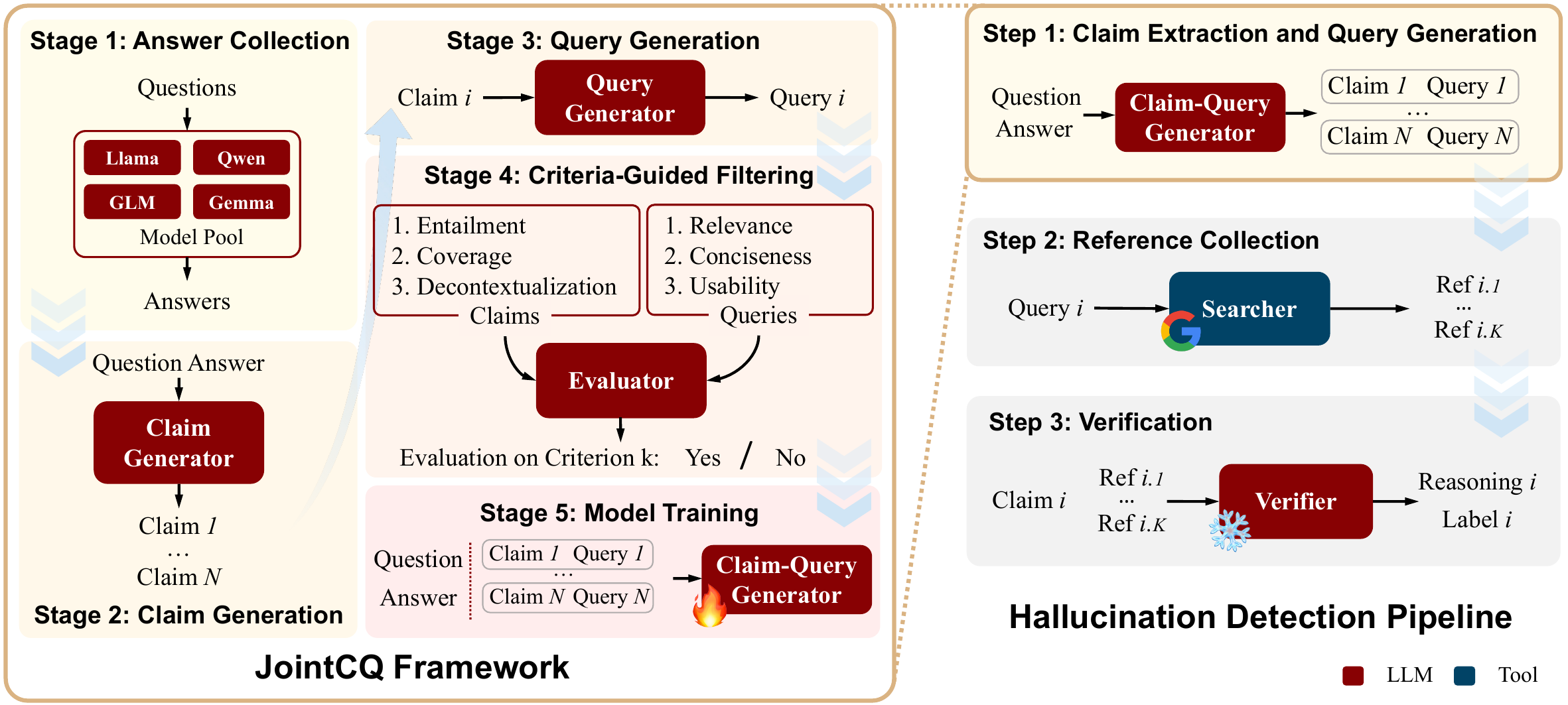}
\caption{Overview of the JointCQ framework (left) and hallucination detection pipeline (right). The claim-query generator is built with the JointCQ framework and can jointly generate claims and their corresponding queries in a single inference step.}
\label{fig:pipeline}
\end{figure*}

To effectively detect factual hallucinations in language model outputs, it is essential to first generate grounded claims along with their corresponding retrieval-oriented queries. This relies a model trained on high-quality and well-aligned claim-query pairs. Therefore, we propose \textbf{JointCQ}, a comprehensive framework that includes both the construction of training data and the training of a joint claim-query generation model. The framework first uses an LLM to generate candidate claims and queries, then applies a rigorous filtering process to ensure data quality. The resulting filtered data is used to finetune a language model that can produce reliable claims and the corresponding queries in a single inference step.

The core strength of JointCQ lies in its criteria-guided data filtering process. Rather than relying on loosely aligned or noisy data, we apply a dual evaluation procedure that filters claims and queries independently. For claims, we assess entailment, coverage, and decontextualization. For queries, we evaluate relevance, conciseness, and usability to ensure that they support effective retrieval and align closely with the associated claims. As a result, the JointCQ framework ensures high-quality training data and enables a more effective joint claim-query generator. This generator serves as a solid foundation for downstream hallucination detection process. Additionally, our framework is fully built upon open-source models and supports both English and Chinese. Experiments on open-domain QA hallucination detection benchmarks demonstrate that our method outperforms strong baselines on both languages, advancing the development of more trustworthy and transparent language model systems.

To summarize, our main contributions are:

\begin{enumerate}[itemsep=1pt, topsep=1pt, leftmargin=*]
    \item We propose \textbf{JointCQ}, a framework that can train a model capable of generating both factual claims and their corresponding search queries in a single inference for factual hallucination detection. The framework is fully built on open-source models, ensuring low cost, high accessibility, and ease of deployment.
    \item We design a dual-stage, criteria-guided filtering strategy to construct high-quality training data in JointCQ, ensuring the model is trained on accurate and well-aligned claim-query pairs.
    \item Experimental results on multiple open-domain QA hallucination detection benchmarks demonstrate that JointCQ substantially improves the factual hallucination detection performance, surpassing several strong baselines.
\end{enumerate}

\section{Hallucination Detection Task}

\subsection{Task Formulation}
Given a question and a corresponding answer generated by a language model, our goal is to detect factual hallucinations at the claim level. We adopt the definition of a factual claim from \citet{ni-etal-2024-afacta}, where a claim is a statement explicitly presenting verifiable facts. Here, a fact is an assertion that can be objectively verified as true or false based on empirical evidence or reality. This claim-level formulation allows for fine-grained hallucination detection. It also supports more targeted verification and modular processing. 

Formally, the task can be described as:
\begin{itemize}[itemsep=1pt, topsep=1pt, leftmargin=*]
    \item \textbf{Input}: A natural language question $q$ and a model-generated answer $a$ that may contain correct information, hallucinations, or unverifiable content.
    \item \textbf{Output}: 
    A set of factual claims $\{c_1, c_2, \dots, c_N\}$ extracted from $(q, a)$, where each claim $c_i$ is assigned with a factuality label $l_i \in \{Correct, Hallucinated,$ $ Unverifiable\}$ indicating its status based on external evidence.
\end{itemize}

\subsection{Pipeline Components}
A standard hallucination detection pipeline typically consists of four sequential steps~\cite{min-etal-2023-factscore,chern2023factool,fatahi-bayat-etal-2023-fleek,wei2024long_safe,cheng-etal-2024-small}: (1) response decomposition, (2) query generation\footnote{Some approaches simplify this step by extracting keywords or directly reusing decomposed segments as queries.}, (3) evidence retrieval, and (4) factual verification. However, this pipeline design often leads to issues such as missing factual details, loss of context, and insufficiently targeted queries.

To address these issues, we redesign the pipeline by unifying the first two stages into a single step using our proposed \textbf{JointCQ} framework. As shown in the right part of Figure~\ref{fig:pipeline}, given a question and its answer, the claim-query generator jointly extracts factual claims and generates corresponding queries. The searcher sends these queries to Google Search via the Serper API\footnote{\url{https://serper.dev}} and retrieves the top-10 snippets as evidence.
Finally, a verifier implemented with \texttt{Qwen3-14B}\footnote{Other LLMs, especially larger models, will work well or even better in this step, but for cost and efficiency consideration, we simply use Qwen3-14B here.}, assesses each claim's factuality against the retrieved snippets.
Appendix \ref{app:impl_pipeline} provides additional information on the implementation of hallucination detection pipeline.

\section{JointCQ Framework}

\subsection{Overview}
This section presents the JointCQ framework, designed to enhance hallucination detection by optimizing the claim extraction and query generation stage (Figure \ref{fig:pipeline}). Central to our approach is the construction of high-quality, well-aligned claim-query training data through a rigorous, criteria-guided filtering process, ensuring effective and efficient supervision. The filtered data is then used to train a joint claim-query generation model capable of producing claim-query pairs in a single inference step.

\subsection{Data Synthesis}
\label{sec:data_synthesis}
\subsubsection{Data Sourcing}
\label{sec:data_sourcing}
The \texttt{question} segment of the ANAH-v2 dataset~\cite{gu2024anah} serves as the core data source. This dataset consists of questions and reference documents, but does not include hallucination labels. We leverage a diverse set of mainstream large language models to generate corresponding answers: \texttt{Qwen2.5-7B-Instruct}~\cite{qwen2025qwen25technicalreport}, \texttt{Llama-3.1-8B-Instruct}~\cite{grattafiori2024llama3herdmodels}, \texttt{gemma-3-4b-it}~\cite{gemmateam2025gemma3technicalreport} and \texttt{glm-4-9b-chat}~\cite{glm2024chatglmfamilylargelanguage}. This ensures the richness of answer variations, thereby laying a comprehensive foundation for extracting diverse factual claims. Consequently, this stage yields a collection of question–answer pairs that serve as input for subsequent stages of supervised data construction.

\subsubsection{Claim Synthesis}
Claim extraction is performed using a 3-shot prompting strategy to guide the claim generation model, \texttt{Qwen3-32B}~\cite{yang2025qwen3}. 
In-context examples are constructed from the same dataset described in the previous section, with output segments manually written.
For each QA pair, we first retrieve the top-3 examples with the highest semantic similarity (measured by the \texttt{paraphrase-multilingual-mpnet-base-v2} embedding model~\cite{reimers-2019-sentence-bert}) and the top-3 examples with the most similar answer length. From this candidate pool of up to six examples, we randomly sample three as the final in-context examples.

The model is instructed to generate clear, factual, and self-contained claims, excluding subjective or ambiguous content. By applying this prompting process, we extract a set of factual claims $\{c_1, \dots, c_N\}$ from each QA pair.

\subsubsection{Query Synthesis}
Query generation adopts a 3-shot prompting strategy, selecting three random examples. The query generator is implemented with \texttt{Qwen3-32B} as well. For each claim $c_i$, a search query $q_i$ is generated, bridging the gap between extracted claims and the evidence retrieval stage. For more details on the data synthesis implementation, please refer to Appendix \ref{app:data_generation}.

\subsection{Criteria-Guided Filtering}
To improve the quality of claims and queries in our training dataset, we use a filtering process on both elements.
The process guarantees that each claim is grounded in input QA pairs and clearly stated, while each query is effective for finding relevant information. Examples of passed and failed claims and queries on each criterion are shown in Table \ref{tab:criteria_examples} in Appendix \ref{app:data_filtering}.

\subsubsection{Claim Evaluation Criteria}
For the selection of claims, we adopt and modify the criteria mentioned by \citet{metropolitansky-larson-2025-towards}:
\begin{itemize}[itemsep=1pt, topsep=1pt, leftmargin=*]
    \item \textit{\textbf{Entailment}: The content of the claims should be fully supported by the source text, i,e, the question and answer.} \\
    Unlike settings where claims are derived solely from answers, we treat the question as an essential part of the context. This is because many answers are underspecified on their own, and only make complete sense when interpreted alongside the question.
    \item \textit{\textbf{Coverage}: The extracted claims should capture all the verifiable factual information in the source text.} \\
    This helps avoid selective reporting or omission of fact-related information.
    \item \textit{\textbf{Decontextualization}: The claim should be understandable on its own, without requiring additional context.} \\
    This criterion follows principles from sentence decontextualization research~\cite{choi-etal-2021-decontextualization}, which emphasize the portability and semantic completeness of isolated textual statements.
\end{itemize}

While grounded in similar theoretical foundations, our use case and filtering process differs from the evaluation framework of \citet{metropolitansky-larson-2025-towards}, where claims are directly used as search queries to retrieve supporting evidence.
We introduce an additional step by generating a separate query for each claim. This query is optimized for external information retrieval (e.g., from a search engine) and is evaluated using its own set of criteria. This distinction is important: it allows us to maintain the factual clarity and independence of each claim while tailoring the retrieval process through purpose-built, query-specific formulations. By separating claim construction from query design, we are able to better control for both the verifiability of the content and the effectiveness of the retrieval process. This separation lead to a total different definition of decontextualization.

\subsubsection{Query Evaluation Criteria}
Unlike claims, query evaluation emphasizes retrieval effectiveness and search-oriented design. Our formulation of query criteria draws from information retrieval theory~\cite{schutze2008introduction, cronen2002predicting}. The criteria are as follows:
\begin{itemize}[itemsep=1pt, topsep=1pt, leftmargin=*]
    \item \textit{\textbf{Relevance}: The query directly relates to the claim, addressing its content, implications, or underlying assumptions.} \\
    This criterion ensures that retrieved information is semantically aligned with the claim, thereby reducing the inclusion of off-topic or tangential evidence. It serves as a basic but essential filter for maintaining consistency between the claim and external knowledge sources.
    \item \textit{\textbf{Conciseness}: The query should be clear and focused on the core information. Avoid multiple complex ideas or detailed descriptions in one query.}\\
    This criterion corresponds to the query clarity principle in IR literature, where shorter and clearer queries can yield more relevant results.
    \item \textit{\textbf{Usability}: The query should use natural, fluent, and easily readable language that can yield relevant and accurate results from Google Search.}\\
    This criterion captures the practical need for queries to be interpretable by real-world search engines. Natural-sounding queries are more likely to elicit high-quality results, both in human-centered and automated search scenarios.
\end{itemize}

\subsubsection{Evaluation Protocol Design}
To implement the filtering at scale, we design a hybrid evaluation protocol that leverages the capabilities of the \texttt{Qwen3-32B} language model. We separate the evaluation procedures for different criteria to minimize cross-dimensional interference and maximize reliability.

For entailment and coverage, we conduct evaluation in a batch-oriented manner, where each batch corresponds to the full set of claims extracted from a single QA pair. This provides the model with sufficient context. 

By contrast, decontextualization is evaluated at the individual claim level, with each claim presented to the model in isolation, absent accompanying claims. This setup directly tests whether the claim remains semantically self-sufficient.

Similarly, evaluation of queries is conducted on an individual basis, with each query-claim pair assessed separately. This ensures a localized evaluation of query quality, unimpeded by interactions with other queries or external context. Appendix \ref{app:data_filtering} offers a more thorough description of the criteria-guided filtering implementation.

\subsection{Model Training}
\subsubsection{Data Preparation}
To mitigate bias toward a specific claim count of each QA pair, we stratify samples by their claim count and enforce per-group sampling limits. 
After stratified sampling, random selection fills remaining quotas, producing a final dataset of 1,000 samples for each language with moderately balanced claim count distributions. We partition each language subset into training and test sets (9:1 ratio), resulting in 1,800 training and 200 validation samples.

\subsubsection{Training Details}
We fine-tune the \texttt{Qwen2.5-14B-} {Instruct}~\cite{qwen2025qwen25technicalreport} model as our Claim-Query Generator, leveraging its strong instruction-following aptitude and computational efficiency for this task. Training runs for 1 epoch on synthetic (claim, query) pairs with a batch size of 128, optimized for memory efficiency on 4$\times$NVIDIA H100 GPUs (80GB VRAM) using DeepSpeed Zero-3 for distributed training. Hyperparameters include a 1e-5 learning rate (10\% linear warmup), and bfloat16 mixed-precision training with gradient checkpointing.

\section{Experiment Setup}
\subsection{Test Sets}
We evaluate our method on two publicly available benchmark datasets across different domains and languages:
\begin{itemize}[itemsep=1pt, topsep=1pt, leftmargin=*]
    \item \textbf{ANAH}~\cite{ji-etal-2024-anah}\footnote{ANAH is a totally different dataset from the ANAH-v2 mentioned in Section \ref{sec:data_sourcing}.}: A bilingual dataset with sentence-level hallucination annotations from LLM responses. We sample 500 QA pairs per language for a 1,000-sample test set supporting both response- and sentence-level evaluation. This size is relatively large compared to similar prior works~\cite{chern2023factool,cheng-etal-2024-small}, allowing for reliable assessment.
    \item \textbf{HalluQA}~\cite{cheng2023halluqa}: A Chinese hallucination detection benchmark for QA task with binary, response-level labels. We use all the 206 fact-related samples for our experiments, following the setup in HaluAgent~\cite{cheng-etal-2024-small}.
\end{itemize}
These test sets cover both English and Chinese, and support multi-granularity hallucination analysis, providing a comprehensive benchmark for evaluating the generalization and robustness of hallucination detection methods.

\subsection{Baselines}
We compare our framework with several strong base LLMs and hallucination detection methods:

\begin{itemize}[itemsep=1pt, topsep=1pt, leftmargin=*]

\item \textbf{GPT-4.1 and DeepSeek R1}~\cite{openai2024gpt4technicalreport,deepseekai2025deepseekr1incentivizingreasoningcapability}: Strong general large language models with competitive capabilities, including hallucination detection ability.

\item \textbf{SelfCheckGPT}~\cite{manakul-etal-2023-selfcheckgpt}: A classical hallucination detection method that detects hallucinations by generating multiple responses from a language model and checking for consistency across them.

\item \textbf{FacTool}~\cite{chern2023factool}: A tool-augmented framework designed for factual error detection across diverse generative tasks.
    
\item \textbf{HaluAgent}~\cite{cheng-etal-2024-small}: An autonomous hallucination detection framework built on small open-source models, integrating multiple tools for fact-checking.
\end{itemize}

\subsection{Evaluation Metrics}
We use \textbf{Accuracy} and hallucination \textbf{F1 score} for both sentence- and response-level evaluation. Unverifiable or failed samples are treated as no hallucination, similar to the setup in FacTool~\cite{chern2023factool}. Evaluation results for only the verifiable samples are in Appendix \ref{app:app_experiments}.

For sentence-level evaluation, claim $c_j$ is aligned to response sentence $s_i$ when: 
(1) $s_i$ is most semantically similar to $c_j$, and (2) cosine similarity\footnote{Texts are embedded with \texttt{paraphrase-multilingual- mpnet-base-v2}~\cite{reimers-2019-sentence-bert}.} exceeds threshold $\theta=0.5$ \footnote{The threshold is empirically chosen to filter out pairs with low semantic relatedness, as text pairs with cosine similarity below 0.5 are typically considered non-matching in semantic similarity tasks.}. 

Let $R$ denote the set of sentences in a response. The aligned claims for $s_i$ are defined as:
\begin{align*}
    C(s_i) = \{ c_j \mid s_i = \arg\max_{s_k \in R} \text{sim}(s_k, c_j) \land \\ \text{sim}(s_i, c_j) \geq \theta \}.
\end{align*}

Hallucination labels are aggregated hierarchically:
\begin{align*}
    H(s_i) &= \mathbb{I}[\exists c_j \in C(s_i): h(c_j)=1],\\
    H(r) &= \mathbb{I}[\exists s_i \in R: H(s_i)=1],
\end{align*}

where $\mathbb{I}[\cdot]$ is the indicator function. This ensures consistent evaluation across annotation granularities. Further details about the experiment setup and results can be found in Appendix \ref{app:app_experiments}.

\begin{table*}[h] 
\small
\centering
\setlength{\tabcolsep}{1.7mm}
\begin{tabular}{lcccccccccccc}
\toprule
\multicolumn{1}{c}{} & \multicolumn{3}{c}{ANAH-en} & \multicolumn{3}{c}{ANAH-zh} & \multicolumn{3}{c}{ANAH-overall} & \multicolumn{3}{c}{HalluQA} \\
 & Acc & F1 & N unv. & Acc & F1 & N unv. & Acc & F1 & N unv. & Acc & F1 & N unv. \\
\midrule
DeepSeek R1 & 61.40 & 42.73 & - & 61.40 & 58.13 & - & 61.40 & 51.63 & - & 76.70 & 74.19 & -\\
GPT-4.1 & 71.80 & 65.01 & - & 61.40 & 56.43 & - & 66.60 & 60.52 & - & 72.82 & 70.53 & - \\
\midrule
SelfCheckGPT & 70.20 & 74.35 & -  & 67.60 & 75.89 & -  & 69.80 & 75.18 & - & 56.31 & 68.97 & - \\
FacTool & 74.20 & \textbf{77.33} & 13 & 68.60 & 76.46 & 11 & 71.40 & 76.86 & 24 & 56.80 & 46.71 & 12 \\
HaluAgent-13B & 72.80 & 70.82 & 21 & 67.20 & 67.97 & 29 & 70.00 & 69.30 & 50 & 78.16* & \textbf{83.75}* & - \\
\midrule
Ours & \textbf{75.80} & 76.95 & 5 & \textbf{72.60} & \textbf{77.58} & 11 & \textbf{74.20} & \textbf{77.29} & 16 & \textbf{80.58} & 83.05 & 5 \\
\bottomrule
\end{tabular}
\caption{Response-level evaluation results. Acc and F1 values are reported in percentage. The results of HaluAgent-13B on HalluQA dataset comes from the paper~\cite{cheng-etal-2024-small}. ``N unv." denotes the number of unverifiable samples.}
\label{tab:result_response-level_correct}
\end{table*}

\section{Results and Analysis}
\begin{table}[h]
\small
\centering
\setlength\tabcolsep{1.05mm}
\begin{tabular}{lcccccc}
\toprule
\multicolumn{1}{c}{} & \multicolumn{2}{c}{ANAH-en} & \multicolumn{2}{c}{ANAH-zh} & \multicolumn{2}{c}{ANAH-all} \\

 & Acc & F1 & Acc & F1 & Acc & F1 \\
\midrule
FacTool & 74.64 & 67.57 & 68.18 & 68.02 & 71.75 & 67.80 \\
SelfCheckGPT & 74.32 & 69.57 & 67.34 & 67.84 & 71.24 & 68.72 \\
\midrule
Ours & \textbf{80.14} & \textbf{70.99} & \textbf{76.16} & \textbf{71.10} & \textbf{78.36} & \textbf{71.04} \\
\hfill w/o filtering & 77.63 & 67.32 & 74.85 & 69.55 & 76.39 & 68.42 \\
\hfill filter $c$ only & 78.59 & 68.91 & 73.54 & 67.35 & 76.33 & 68.15 \\
\hfill filter $q$ only & 78.88 & 68.85 & 75.09 & 68.95 & 77.19 & 68.90 \\
\hfill w/o $q$ & 77.63 & 66.38 & 73.18 & 66.06 & 75.64 & 66.22 \\
\hfill replace CQG & 75.89 & 65.28 & 73.42 & 67.73 & 74.79 & 66.48 \\
\bottomrule
\end{tabular}
\caption{Sentence-level evaluation of hallucination detection on ANAH dataset.}
\label{tab:result_sentence-level}
\end{table}

\subsection{Main Results}
Table \ref{tab:result_response-level_correct} presents the response-level evaluation results.
Our method achieved competitive results, with the highest accuracy scores on ANAH-overall (74.20\%) and HalluQA (80.58\%). FacTool showed lower accuracy on HalluQA but performed moderately on ANAH. 
While HaluAgent-13B achieved high accuracy on ANAH-en and HalluQA, its performance dropped significantly on ANAH-zh, suggesting language- and domain-dependent limitations.Our method also resulted in the fewest unverifiable samples and exhibited better usability.

Table \ref{tab:result_sentence-level} presents sentence-level hallucination detection results on the ANAH dataset. Our method achieves state-of-the-art performance across all settings, attaining the highest scores in both English (ANAH-en: 80.14\% Acc/70.99\% F1) and Chinese (ANAH-zh: 76.16\% Acc/71.10\% F1) verifiable samples, with consistent advantages of +5\textasciitilde8\% accuracy and +3\textasciitilde4 F1 points over FacTool. 

Overall, the experimental results demonstrate that our proposed framework outperforms the baseline methods in most cases, whether evaluating at the response level or the sentence level. Our framework shows better accuracy and F1 scores, indicating its strong capability in detecting factual hallucinations on the open-domain QA task.

\subsection{Necessity of Queries}
Previous work~\cite{metropolitansky-larson-2025-towards} state that claims are used to retrieve relevant information from sources, which is different from our settings of using additional queries. To assess the importance of the query generation step, we conduct an ablation study where the generated queries are replaced with claims, while keeping all other components unchanged.
The experimental results are presented in Table~\ref{tab:result_sentence-level}, indicated as ``w/o $q$". Compared to the complete implementation, performance drops noticeably for both Chinese and English, with a decline of 4.82 points in overall hallucination F1 score. These results underscore the necessity of incorporating a dedicated query generation step. Notably, our framework integrates claim extraction and query generation within a single inference pass, introducing minimal additional computational cost.

\subsection{Effectiveness of Criteria-guided Filtering}
To evaluate the impact of criteria-guided filtering, we compare three experimental settings: (1) no filtering applied to either claims or queries (w/o filtering), (2) filtering applied only to claims (filter $c$ only), and (3) filtering applied only to queries (filter $q$ only). The training data size and sampling strategies remain consistent with the main experiment. As shown in Table \ref{tab:result_sentence-level}, omitting filtering in any configuration results in a performance decline, though the magnitude varies. This demonstrates that our curated filtering criteria enhance the quality of both claims and queries, leading to improved hallucination detection performance.

\subsection{Effectiveness of Claim-Query Generator}
We conduct an additional ablation study by replacing the Claim-Query Generator with the separate claim synthesis and query synthesis steps with base LLMs described in Section~\ref{sec:data_synthesis}, while keeping the rest of pipeline the same. The results, shown in Table~\ref{tab:result_sentence-level} under the setting ``replace CQG", indicate a clear drop in performance compared to the full JointCQ framework. Notably, even when compared to earlier ablations on criteria-guided filtering, the base synthesis approach performs worse. These findings highlight the advantage of jointly generating claims and queries in a single model inference, and further demonstrate the effectiveness of the JointCQ framework.

\subsection{Reliability of Verifier}
\label{sec:reliability_of_verifier}
To evaluate the reliability of the verifier, we randomly sample 50 claims per language, along with their corresponding search results. Each claim is manually annotated as Correct, Hallucinated, or Unverifiable based on the retrieved evidence. More details about manual annotation are presented in Appendix \ref{app:annotation}. Among the 93 claims labeled as verifiable, the model verifier \texttt{Qwen3-14B} achieves a consistency rate of 91.40\% with human annotations. This result indicates that current large language models perform well on the verification task. The bottleneck in hallucination detection performance, therefore, lies in earlier stages, supporting our initial motivation. By focusing on generating higher-quality claims and queries, the proposed JointCQ framework contributes to improved detection accuracy.

\subsection{Efficiency Analysis}
\label{efficiency_analysis}
\begin{table}[H]
\centering
\small
\begin{tabular}{lcc}
\toprule
\textbf{Method} & \textbf{Search / jud.} & \textbf{Inference / sample} \\
\midrule
FacTool        & 2  & 13.40 \\
HaluAgent       & 1.38   & 5.24  \\
Ours            & 1   & 4.93  \\
\bottomrule
\end{tabular}
\caption{Average search call per judgement and inference call per QA sample. Here judgement refers to a decision of whether the given text segment contains hallucination.}
\label{tab:efficiency}
\end{table}
We evaluate the efficiency of the hallucination detection pipeline on 200 QA examples from the ANAH dataset. The end-to-end processing takes 599 seconds on a server with 4 NVIDIA H100 GPUs using the \texttt{vllm} engine. The main bottleneck is the reference search stage (303s), while inference remains efficient.

As shown in Table~\ref{tab:efficiency}, our framework requires only 1 search API call per judgement and 4.93 model inferences per sample, significantly fewer than \textbf{FacTool} and comparable to \textbf{HaluAgent}. Unlike HaluAgent, which produces coarse response-level labels, JointCQ performs fine-grained, claim-level hallucination detection.
In addition, while both FacTool and HaluAgent rely on APIs of closed-source models, our framework is built entirely on open-source models, offering greater accessibility and lower deployment cost.

\subsection{Case Study}
\begin{figure}[ht]
\centering
\includegraphics[width=1.0\linewidth]{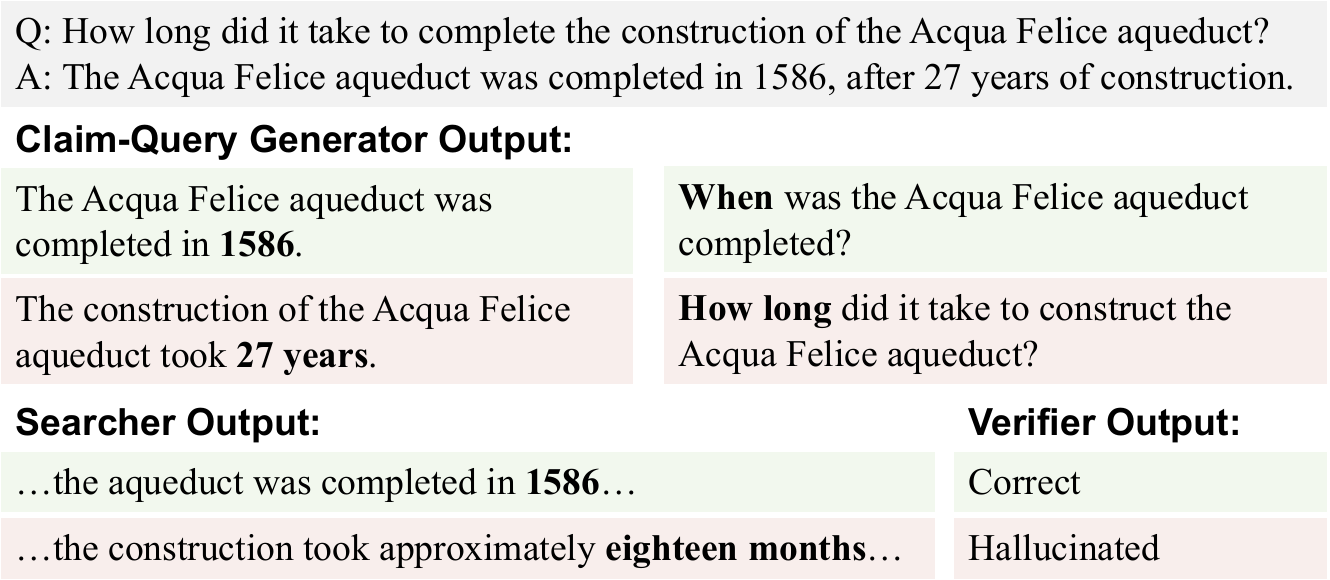}
\caption{An example of the detection process.}
\label{fig:case_study}
\end{figure}

To illustrate the effectiveness of our framework, we present an example in Figure \ref{fig:case_study}. This case illustrates two key observations. First, claims are typically more fine-grained than full sentences. Instead of assessing the entire sentence, breaking it into individual claims enables more precise identification of hallucinated content.
Second, the queries are closely aligned with the specific elements of each claim, targeting the parts most likely to be incorrect. Here, the queries focus on the year of completion and the period of construction. This targeted querying improves retrieval relevance.

\section{Related Work}
\subsection{Factual Hallucination Detection with Web Search or Retrieval}

A prominent line of research enhances factuality detection using external knowledge sources in a ``retrieve-and-verify'' paradigm, often decomposing content into factual units for fine-grained analysis. \citet{min-etal-2023-factscore} propose FActScore, which verifies atomic facts against Wikipedia, offering interpretability but limited by a single-source knowledge base and explicit entity requirements. \citet{chern2023factool} introduce FacTool, a unified framework across tasks such as QA, code generation, and math, while FLEEK~\cite{fatahi-bayat-etal-2023-fleek} incorporates both detection and correction. \citet{qin2025don} propose a retrieval-augmented framework that proactively verifies false premises in queries before generation, related to our claim–query paradigm but focused on pre-generation validation. Agent-based approaches with more flexibility include SAFE~\cite{wei2024long_safe} and HaluAgent~\cite{cheng-etal-2024-small}, and KnowHalu~\cite{zhang2024knowhalu} introduces a two-phase, multi-form knowledge framework with stepwise reasoning for structured factual verification.

The most closely related to our work are FacTool~\cite{chern2023factool} and HaluAgent~\cite{cheng-etal-2024-small}. While FacTool provides a general framework across tasks, it incurs high computational cost as shown in Section \ref{efficiency_analysis}. HaluAgent adopts a more flexible agent-based approach, but it operates primarily at the response level and lacks fine-grained control over hallucination localization. In contrast, our method enables efficient, fine-grained hallucination detection.

\subsection{Claim Extraction and Claim-Level Fact Checking}
Claim extraction enables fine-grained factuality assessment by isolating verifiable statements. FEVERFact~\cite{ullrich2025claim} provides a benchmark evaluating atomicity, fluency, and faithfulness.
\citet{metropolitansky-larson-2025-towards} introduces Claimify, an LLM-based method that extracts claims only when confident in interpretation. The paper also proposes a standardized framework to assess extraction quality in terms of coverage and decontextualization. We designed the training data filtering step based on the criteria introduced in this work.
AFaCTA~\cite{ni-etal-2024-afacta} leverages LLMs for consistent claim annotation, producing the PoliClaim dataset. HalluMeasure~\cite{akbar-etal-2024-hallumeasure} decomposes LLM outputs into atomic claims and detects hallucinations via Chain-of-Thought reasoning. However, its applicability is limited to summarization tasks and it lacks a retrieval component suited for addressing factual hallucinations.
FactSelfCheck~\cite{sawczyn2025factselfcheck} uses a black-box, sampling-based fact-level approach with knowledge-graph triples to enable precise claim-level detection and correction without external resources, complementing retrieval- and reasoning-based methods.

\subsection{Efficient Hallucination Detection Methods}
Another type of approaches aims to detect hallucinations without relying on external knowledge, prioritizing efficiency.
SelfCheckGPT~\cite{manakul-etal-2023-selfcheckgpt} proposes a zero-resource, black-box method that assesses hallucination by measuring the consistency between multiple sampled outputs using metrics such as BERTScore, NLI inference, and QA agreement.
To address the overconfidence or underconfidence of model-internal probabilities, \citet{zhang-etal-2023-enhancing-uncertainty} introduce an uncertainty-based method using a proxy model to adjust token-level probabilities based on contextual informativeness and reliability. 
HaloCheck~\cite{elaraby2023halo} evaluates hallucination in weaker open-source LLMs through consistency judgments among multiple responses using an NLI model.
While these approaches incur low computational cost and avoid reliance on external resources, their reliability for factual verification remains limited, as they depend on internal uncertainty signals rather than grounded world knowledge.

\section{Conclusion}
In this work, we designed a three-stage pipeline (claim-query generation, evidence retrieval, and verification) for factual hallucination detection and introduced JointCQ, a framework that produces high-quality claims and queries to build a reliable claim-query generator. Unlike prior methods that depend on closed-source APIs, our framework is fully based on open-source models and supports both English and Chinese, making it easily accessible and broadly applicable. Experimental results demonstrate that JointCQ achieves strongest performance over multiple benchmarks, marking a step forward in building more trustworthy and transparent language model systems.

\section*{Limitations}
Despite the promising results of our framework, several limitations should be noted. First, the pipeline is primarily designed for general open-domain QA tasks. While QA represents a fundamental and broadly applicable task format, extending the framework to other NLP tasks would require additional adaptation and validation. Second, our evidence retrieval component relies on Google Search, which exposes the system to the inherent limitations of the search engine. Nevertheless, leveraging such external services remains one of the most effective approaches for obtaining up-to-date and reliable information, and this strategy is commonly adopted in contemporary hallucination detection studies.

\bibliography{jointcq_custom}

\appendix
\begin{CJK}{UTF8}{gkai}

\section{Implementation of the JointCQ Framework}
\label{app:impl_framework}
\subsection{Data Generation}
\label{app:data_generation}
We sample 2,000 Chinese and 2,000 English questions from the ANAH-v2~\cite{gu2024anah} dataset. Answers are generated by four LLMs: \texttt{Qwen2.5-7B-Instruct}~\cite{qwen2025qwen25technicalreport}, \texttt{Llama-3.1-8B-Instruct}~\cite{grattafiori2024llama3herdmodels}, \texttt{gemma-3-4b-it}~\cite{gemmateam2025gemma3technicalreport}, \texttt{glm-9b-chat}~\cite{glm2024chatglmfamilylargelanguage}. The prompt consists of only the question (without additional instructions) to simulate real-world usage. Detailed statistics are provided in Table \ref{tab:stat_data_sourcing}.

\begin{table}[H]
    \centering
    \small
    \begin{tabular}{lccccc}
    \toprule
    Model     & Qwen & Llama & Gemma & GLM & Total \\
    \midrule
    N en     & 495 & 490 & 502 & 513 & 2000\\
    N zh     & 646 & 0 & 687 & 667 & 2000\\
    \midrule
    Total      & 1141 & 490 & 1189 & 1180 & 4000\\
    \bottomrule
    \end{tabular}
    \caption{Statistics of generated answers in data sourcing stage.}
    \label{tab:stat_data_sourcing}
\end{table}

We then synthesize claims and queries for QA pairs using few-shot prompting. The claim generation prompt is provided in Tables \ref{tab:en_prompt_claim_generation} and \ref{tab:zh_prompt_claim_generation}, while the query generation prompt is detailed in Tables \ref{tab:en_prompt_query_generation} and \ref{tab:zh_prompt_query_generation}. The generator's temperature is set to 0.9.

\begin{table}[H]
\setlength{\tabcolsep}{1.0mm}
\centering
\small
\begin{tabular}{lcccccc}
    \toprule
     & \multicolumn{3}{c}{Claim} & \multicolumn{3}{c}{Query}\\
    Criterion     & Ent. & Cov. & Dec. & Rel. & Con. & Usa. \\
    \midrule
    Pass cnt     &  3,843 & 3,635 & 2,472 & 29,216 & 28,924 & 29,258\\
    Pass rate(\%)     & 96.08 & 90.88 & 61.80 & 99.23 & 98.23 & 99.37\\
    \bottomrule
    \end{tabular}
\caption{Statistics of data filtering.}
\label{tab:statistics_filtering}
\end{table}

\subsection{Data Filtering}
\label{app:data_filtering}
Prompt templates for claim and query filtering are shown in Tables \ref{tab:en_prompt_claim_filtering} and \ref{tab:zh_prompt_claim_filtering} (claims) and Tables \ref{tab:en_prompt_query_filtering} and \ref{tab:zh_prompt_query_filtering} (queries). Each evaluation assesses only one criterion at a time, with the evaluator's temperature set to 0.0 for maximum accuracy.

Initial filtering statistics (Table \ref{tab:statistics_filtering}) reveal that decontextualization is the most challenging criterion, with an initial pass rate of 61.8\%, while other criteria maintain pass rates above 90\%. For samples failing the initial filter, we iteratively repeat the synthesis and filtering process until obtaining over 3,000 qualified samples for subsequent training data sampling.

\section{Implementation of Hallucination Detection Pipeline}
\label{app:impl_pipeline}
The claim and query generation process uses the prompt templates shown in Tables \ref{tab:en_prompt_cqg} and \ref{tab:zh_prompt_cqg}. During the search stage, we configure the system to return 10 results per query. For verification, we employ the prompt templates in Tables \ref{tab:en_prompt_verifier} and \ref{tab:zh_prompt_verifer}. The same model generates outputs for both languages, with only the prompt templates differing. During postprocessing, responses labeled ``Irrelevant" are automatically mapped to ``Unverifiable". To minimize the influence of randomness, the temperature parameters of the model are uniformly set to 0.

\section{Experiments}
\label{app:app_experiments}
\subsection{Implementation of Baselines}
\label{app:impl_baseline}
We employ LLMs as baseline for our response-level evaluation. The hallucination detection prompts for these LLMs are provided in Tables \ref{tab:en_prompt_baseline} and \ref{tab:zh_prompt_baseline}, supporting only binary classification at the response level.

We configure SelfCheckGPT~\cite{manakul-etal-2023-selfcheckgpt} with a sample size of 20 and temperature of 1.0, computing consistency scores using the recommended NLI method.

For HaluAgent~\cite{cheng-etal-2024-small} and FacTool~\cite{chern2023factool}, we utilize GPT-4.1~\cite{openai2024gpt4technicalreport} through the GPT API for all external model calls and keep other inference parameters.

\subsection{Results of Different Evaluation Settings}
\label{app:results_eval}
We propose an alternative evaluation approach that excludes unverifiable or failed samples, focusing solely on the verifiable portions. Notably, the composition of verifiable samples varies across different evaluation methods.

Response-level evaluation results are presented in Table \ref{tab:app_result_response-level}. Sentence-level evaluation results are shown in Table \ref{tab:app_result_sentence-level}.

Our method demonstrates consistent superiority over baseline approaches across both evaluation settings, maintaining robust performance.

\section{Supplementary Information on Manual Annotation}
\label{app:annotation}
To assess the reliability of the verifier, we manually annotate a set of claims and compare the verifier model’s predictions against these human-provided labels (Section \ref{sec:reliability_of_verifier}). This section provides additional details about the annotation process.
We recruit three volunteers familiar with the topic of hallucinations in LLMs. Each claim is independently annotated by one annotator. For each annotation, the annotator is provided with the claim and the corresponding retrieved documents. The annotation guidelines are consistent with the evaluation criteria presented in Tables \ref{tab:en_prompt_verifier} and \ref{tab:zh_prompt_verifer}. Annotators are informed that the dataset and the resulting annotations are used solely for research purposes.

\section{AI Usage Disclosure}
In this work, we employ generative AI to support data analysis and enhance our manuscript. While using AI tools, we meticulously evaluate and edit the output to maintain the precision and credibility of our research.

\section{Ethical Considerations}
We carefully consider the ethical aspects of our work on hallucination detection in general-domain question answering. All hallucinated contents in our datasets are explicitly labeled to ensure transparent and responsible use. We expect that the research poses minimal risks, as it does not involve sensitive data or human subjects. Our study uses only publicly available datasets and pretrained models that are licensed for academic use, and our use of these resources strictly follows their intended research purposes. The data we use do not contain any personally identifiable or sensitive information, and we assume that the original dataset providers perform appropriate anonymization and content filtering. The artifacts (datasets and models) developed in this work are released for research purposes only under terms consistent with the original licenses. 

\begin{table*}[h]
\small
\centering
\setlength\tabcolsep{4pt}
\begin{tabularx}{\textwidth}{l X X X}
\toprule
Criterion & Input QA/Claim & Failed & Passed\\
\midrule
Entailment & Q: Where did Judas Iscariot lead the armed guard to arrest Jesus?\par A: In the Garden of Gethsemane. & The Latvian name for the Convent Yard is ``Karmelītes parkāts." & Judas Iscariot led the armed guard to arrest Jesus in the Garden of Gethsemane. \\
\midrule
Coverage & Q: When was the railway line beyond Dennington to Port Fairy closed?\par A: It was closed on 14 December 1982. This closure was part of a broader trend ...
& The railway line from Dennington to Port Fairy in Victoria, Australia, was closed on 14 December 1982. 
& The railway line from Dennington to Port Fairy in Victoria, Australia, was closed on 14 December 1982.\par The closure of ... was part of a broader trend ...
\\
\midrule
Decontext. & 
Q: Which university library ... ?\par
A: ... the Special Collections department has an archive dedicated to De Niro, which includes scripts ... 
& ... \par The archive includes scripts ... & 
... \par The archive dedicated to Robert De Niro in the Special Collections department includes scripts ...\\

\midrule
Relevance & Johann Strauss II is Mozart's father. & Who is Johann Strauss II's father? &  Who is Mozart's father?\\
\midrule
Conciseness & Vines and grapes represent the connection between Christ and the Eucharist, as well as the idea of spiritual growth and abundance. & What is the symbolism of vines and grapes in Christianity, particularly their connection to Christ, the Eucharist, spiritual growth, and abundance? & What is the symbolism of vines and grapes in the Christianity?  \\
\midrule
Usability & Comte asserted that reason is not a source of knowledge but a tool for understanding knowledge obtained through observation. & Auguste Comte reason knowledge source observation assertion & What was Comte's view on the role of reason in acquiring knowledge? \\
\bottomrule
\end{tabularx}
\caption{Passed and failed examples of evaluation criteria. The criteria for claims are entailment, coverage, and decontextualization. The criteria for queries are relevance, conciseness, and usability.}
\label{tab:criteria_examples}
\end{table*}

\begin{table*}[h]
\centering
\small
\begin{tabular}{lcccccccccccc}
\toprule
\multicolumn{1}{c}{} & \multicolumn{3}{c}{ANAH-en} & \multicolumn{3}{c}{ANAH-zh} & \multicolumn{3}{c}{ANAH-overall} & \multicolumn{3}{c}{HalluQA} \\
 & Acc & F1 & N & Acc & F1 & N & Acc & F1 & N & Acc & F1 & N \\
\midrule
FacTool & 74.54 & \textbf{78.01} & 487 & 68.92 & 77.04 & 489 & 71.72 & 77.49 & 976 & 56.20 & 42.18 & 194\\
HaluAgent-13B & 75.99 & 73.44 & 479 & 68.58 & 70.16 & 471 & 72.32 & 71.69 & 950 & 78.16 & 83.75 & -\\
\midrule
Ours & \textbf{76.36} & 77.54 & 495 & \textbf{73.62} & \textbf{78.61} & 489 & \textbf{75.00} & \textbf{78.11} & 984 & \textbf{82.09} & \textbf{84.08} & 201\\
\bottomrule
\end{tabular}
\caption{Response-level evaluation results for the verifiable part. Accuracy (Acc) and F1 scores are reported as percentages. The results for HaluAgent-13B on the HalluQA dataset are sourced from ~\cite{cheng-etal-2024-small}. Here, N denotes the number of samples used for metric calculation: ANAH contains 500 samples per language, while HalluQA consists of 206 samples.}
\label{tab:app_result_response-level}
\end{table*}

\begin{table*}[h]
\small
\centering
\setlength\tabcolsep{4mm}
\begin{tabular}{lccccccccc}
\toprule
\multicolumn{1}{c}{} & \multicolumn{3}{c}{ANAH-en} & \multicolumn{3}{c}{ANAH-zh} & \multicolumn{3}{c}{ANAH-all} \\

 & Acc & F1 & N & Acc & F1 & N & Acc & F1 & N\\
\midrule
FacTool & 74.66 & 69.54 & 947 & 68.26 & 69.27 & 794 & 71.74 & 69.40 & 1741\\
\midrule
Ours &  \textbf{80.77} & \textbf{74.34} & 905 & \textbf{76.63} & \textbf{74.01} & 736 & \textbf{78.92} & \textbf{74.22} & 1641\\
\hfill w/o filtering & 78.16 & 70.81 & 902 & 76.48 & 73.81 & 727 & 77.41 & 72.29 & 1629\\
\hfill filter $c$ only & 79.38 & 72.46 & 907 & 74.21 & 71.01 & 725 & 77.08 & 71.15 & 1632 \\
\hfill filter $q$ only & 79.91 & 72.78 & 901 & 76.38 & 73.07 & 724 & 78.34 & 72.92 & 1625\\
\hfill w/o $q$ & 78.79 & 70.68 & 896 & 73.94 & 69.75 & 729 & 76.62 & 70.22 & 1625\\
\hfill replace CQG & 76.40 & 69.02 & 894 & 75.57 & 73.12 & 704 & 76.03 & 71.01 & 1598\\
\bottomrule
\end{tabular}
\caption{Sentence-level hallucination detection results for the verifiable part of the ANAH dataset. The evaluation covers 1,037 English sentences and 839 Chinese sentences.}
\label{tab:app_result_sentence-level}
\end{table*}

\begin{table*}
\centering
\small
\begin{tabularx}{\textwidth}{X}
\toprule
English Prompt Template of Claim Synthesis\\
\midrule
\#\#\# Task\\
Given a pair of question and answer as input, your task is to extract all claims.\\
\\
\#\#\# Task Rules\\
When extracting claims, strictly follow these rules:\\
1. Claims must be factual statements that can be verified or refuted. Exclude subjective opinions, emotional expressions, and vague judgments.\\
2. Each claim must be semantically complete and independently understandable without relying on context.\\
3. Avoid ambiguous pronouns in claims. Use specific nouns for clarity.\\
4. Extract and output all qualifying claims, with each claim on a separate line.\\
5. If no claims meeting the above criteria exist in the input, output ``No claims."\\
6. Strictly follow the specified format in the response, without adding extra explanations or unrelated content.\\
\\
\#\#\# Examples\\
\{examples\}\\
\\
\#\#\# Input\\
{[Question]}\\
\{question\}\\
{[Answer]}\\
\{answer\}\\
{[Claims]}\\
\bottomrule
\end{tabularx}
\caption{English prompt template of claim synthesis.}
\label{tab:en_prompt_claim_generation}
\end{table*}

\begin{table*}
\centering
\small
\begin{tabularx}{\textwidth}{X}
\toprule
Chinese Prompt Template of Claim Synthesis\\
\midrule
\#\#\# 任务\\
给定一对问题和回答作为输入，你的任务是提取所有的陈述。\\
\\
\#\#\# 任务规则\\
提取陈述时请严格遵循以下规则：\\
1. 陈述必须是可以核实或驳斥的事实性声明。排除主观意见、情绪表达和模糊判断。\\
2. 每条陈述必须语义完整，不依赖上下文即可独立理解其含义。\\
3. 陈述中禁止使用指代不明的代词，必须使用具体名词表述。\\
4. 必须提取并输出所有符合条件的陈述，每条陈述独占一行。\\
5. 当输入中不存在符合上述标准的陈述时，输出``无陈述"。\\
6. 必须严格按照指定格式回复，不得添加其他内容，不得添加多余的解释说明。\\
\\
\#\#\# 示例\\
\{examples\}\\
\\
\#\#\# 输入\\
{[问题]}\\
\{question\}\\
{[回答]}\\
\{answer\}\\
{[陈述]}\\
\bottomrule
\end{tabularx}
\caption{Chinese prompt template of claim synthesis.}
\label{tab:zh_prompt_claim_generation}
\end{table*}

\begin{table*}
\centering
\small
\begin{tabularx}{\textwidth}{X}
\toprule
English Prompt Template of Query Synthesis\\
\midrule
\#\#\# Task\\
Given a claim, your task is to generate a search engine query to help fact-check the claim.\\
\\
\#\#\# Task Rules\\
When generating the query, strictly follow these rules:\\
1. The query should be concise and clear, specifically targeting the claim to be verified.\\
2. The query should be applicable to search engines and can help users obtain valid information.\\
3. Always output a query.\\
4. If there is nothing to query, output ``No query".\\
5. You must strictly follow the specified format. Do not add any extra content or explanations.\\
\\
\#\#\# Examples\\
\{examples\}\\
\\
\#\#\# Input\\
{[Claim]}\\
\{claim\}\\
{[Query]}\\
\bottomrule
\end{tabularx}
\caption{English prompt template of query synthesis.}
\label{tab:en_prompt_query_generation}
\end{table*}

\begin{table*}
\centering
\small
\begin{tabularx}{\textwidth}{X}
\toprule
Chinese Prompt Template of Query Synthesis\\
\midrule
\#\#\# 任务\\
给定一条陈述，你的任务是生成一条搜索引擎查询，用于协助对该陈述进行事实核查。\\
\\
\#\#\# 任务规则\\
生成查询时请严格遵循以下规则：\\
1. 查询应当简洁明确，对待验证的陈述具有针对性。\\
2. 查询能够应用于搜索引擎的搜索，帮助用户获取有效信息。\\
3. 始终输出一条查询语句。\\
4. 若无待查询的内容，直接输出“无查询”。\\
5. 必须严格按照指定格式回复，不得添加其他内容，不得添加多余的解释说明。\\
\\
\#\#\# 示例\\
\{examples\}\\
\\
\#\#\# 输入\\
{[陈述]}\\
\{claim\}\\
{[查询]}\\
\bottomrule
\end{tabularx}
\caption{Chinese prompt template of query synthesis.}
\label{tab:zh_prompt_query_generation}
\end{table*}

\begin{table*}
\centering
\small
\begin{tabularx}{\textwidth}{X}
\toprule
English Prompt Template of Claim Filtering\\
\midrule
\#\#\# Task\\
You are provided with a question, its answer, a set of claims (a claim) extracted from the QA pair. Your task is to assess whether the claim(s) satisfy the specific criterion.\\
\\
\#\#\# Evaluation Criteria\\
The claim(s) should meet the following criterion:\\
Entailment: The content of the claims should be fully supported by the source text. Review each statement point by point to ensure that every statement is fully supported.\\
\texttt{OR}\\
Coverage: The extracted claims should capture all the verifiable factual information in the source text. Evaluate all claims collectively against the question and answer to verify full coverage.\\
\texttt{OR}\\
Decontextualization: The claim should be understandable on its own, without requiring additional context.\\
\\
If the claim(s) meet the criterion, respond with ``Yes"; otherwise, respond with ``No".\\
\\
\#\#\# Input\\
{[Question]}\\
\{question\}\\
{[Answer]}\\
\{answer\}\\
{[Claim(s)]}\\
\{claims\}\\
\bottomrule
\end{tabularx}
\caption{English prompt template of claim filtering.}
\label{tab:en_prompt_claim_filtering}
\end{table*}

\begin{table*}
\centering
\small
\begin{tabularx}{\textwidth}{X}
\toprule
Chinese Prompt Template of Claim Filtering\\
\midrule
\#\#\# 任务\\
给定一个问题、其答案、一组(条)从问答对中提取的陈述，你的任务是评估这些(条)陈述是否满足特定的标准。\\
\\
\#\#\# 评估标准\\
陈述应当满足以下标准：\\
蕴含性：陈述的内容应完全由原文支持。逐条展开检查陈述，确保每条陈述都能被支持。\\
\texttt{OR}\\
覆盖性：提取出的这组陈述应涵盖原文中所有可验证的事实信息。视所有陈述为一个整体并与问答进行比较以确保覆盖性。\\
\texttt{OR}\\
去上下文化：每条陈述应在不需要额外上下文的情况下可以被理解。\\
\\
如果这些(条)陈述符合标准，请回答“是”；否则，请回答“否”。\\
\\
\#\#\# 输入\\
{[问题]}\\
\{question\}\\
{[回答]}\\
\{answer\}\\
{[陈述]}\\
\{claims\}\\
\bottomrule
\end{tabularx}
\caption{Chinese prompt template of claim filtering.}
\label{tab:zh_prompt_claim_filtering}
\end{table*}

\begin{table*}
\centering
\small
\begin{tabularx}{\textwidth}{X}
\toprule
English Prompt Template of Query Filtering\\
\midrule
\#\#\# Task\\
You are given a claim and a query intended for Google Search. Your task is to evaluate whether the query satisfies the specific criterion.\\
\\
\#\#\# Evaluation Criteria\\
The query is considered helpful if it meets the following criterion:\\
Relevance: The query directly relates to the claim, addressing its content, implications, or underlying assumptions.\\
\texttt{OR}\\
Conciseness: The query should be clear and focused on the core information, avoiding multiple complex ideas or detailed descriptions in one query.\\
\texttt{OR}\\
Usability: The query should use natural, fluent, and easily readable language that can yield relevant and accurate results from Google Search.\\
\\
If the query meets the criterion, respond with “Yes”; otherwise, respond with “No”. No additional explanation is allowed.\\
\\
\#\#\# Input\\
{[Claim]}\\
\{claim\}\\
{[Query]}\\
\{query\}\\
\bottomrule
\end{tabularx}
\caption{English prompt template of Query filtering.}
\label{tab:en_prompt_query_filtering}
\end{table*}

\begin{table*}
\centering
\small
\begin{tabularx}{\textwidth}{X}
\toprule
Chinese Prompt Template of Query Filtering\\
\midrule
\#\#\# 任务\\
给定一条陈述和一条用于Google搜索的查询，你的任务是评估该查询是否满足特定的标准。\\
\\
\#\#\# 评估标准\\
如果查询符合以下标准，则认为它是有帮助的：\\
相关性：提问需紧扣陈述本身，涉及其内容、含义或背后的假设。\\
\texttt{OR}\\
简洁性：查询应简明扼要，聚焦核心信息，避免在一个查询中包含多个复杂概念或细节描述。\\
\texttt{OR}\\
可用性：查询应使用自然、流畅且易读的语言，以便从Google搜索中获得相关且准确的结果。\\
\\
如果查询满足以上标准，请回答“是”；否则，请回答“否”。不允许输出任何额外解释。\\
\\
\#\#\# 输入\\
{[陈述]}\\
\{claim\}\\
{[查询]}\\
\{query\}\\
\bottomrule
\end{tabularx}
\caption{Chinese prompt template of query filtering.}
\label{tab:zh_prompt_query_filtering}
\end{table*}

\begin{table*}
\centering
\small
\begin{tabularx}{\textwidth}{X}
\toprule
English Prompt and Response Templates of Claim-Query Generator\\
\midrule
\#\#\# Task\\
Given a question and an answer as input, your task is to extract all claims, and generate a search engine query for each claim to help fact-check the claims.\\
\\
\#\#\# Task Rules\\
When extracting claims, strictly follow these rules:\\
1. Claims must be factual statements that can be verified or refuted. Exclude subjective opinions, emotional expressions, and vague judgments.\\
2. Each claim must be semantically complete and independently understandable without relying on context.\\
3. Avoid ambiguous pronouns in claims. Use specific nouns for clarity.\\
4. Extract and output all qualifying claims, with each claim on a separate line.\\
5. If no claims meeting the above criteria exist in the input, output ``No claims."\\
\\
When generating the queries, strictly follow these rules:\\
1. The queries should be concise and clear, specifically targeting the claims to be verified.\\
2. The queries should be applicable to search engines and can help users obtain valid information.\\
3. If there is nothing to query, output ``No query".\\
\\
\#\#\# Input\\
{[Question]}\\
\{question\}\\
{[Answer]}\\
\{answer\}\\
\midrule
{[Claims]}\\
\{claims\}\\
{[Queries]}\\
\{queries\}\\
{[End]}\\
\bottomrule
\end{tabularx}
\caption{English prompt and response templates of Claim-Query Generator.}
\label{tab:en_prompt_cqg}
\end{table*}

\begin{table*}
\centering
\small
\begin{tabularx}{\textwidth}{X}
\toprule
Chinese Prompt and Response Templates of Claim-Query Generator\\
\midrule
\#\#\# 任务\\
给定问题和回答作为输入，你的任务是提取所有的陈述，然后为每条陈述生成一条搜索引擎查询，用于协助对陈述进行事实核查。\\
\\
\#\#\# 任务规则\\
提取陈述时请严格遵循以下规则：\\
1. 陈述必须是可以核实或驳斥的事实性声明。排除主观意见、情绪表达和模糊判断。\\
2. 每条陈述必须语义完整，不依赖上下文即可独立理解其含义。\\
3. 陈述中禁止使用指代不明的代词，必须使用具体名词表述。\\
4. 必须提取并输出所有符合条件的陈述，每条陈述独占一行。\\
5. 当输入中不存在符合上述标准的陈述时，输出``无陈述"。\\
\\
生成查询时请严格遵循以下规则：\\
1. 查询应当简洁明确，对待验证的陈述具有针对性。\\
2. 查询能够应用于搜索引擎的搜索，帮助用户获取有效信息。\\
3. 若无待查询的内容，直接输出“无查询”。\\
\\
\#\#\# 输入\\
{[问题]}\\
\{question\}\\
{[回答]}\\
\{answer\}\\
\midrule
{[陈述]}\\
\{claims\}\\
{[查询]}\\
\{queries\}\\
{[结束]}\\
\bottomrule
\end{tabularx}
\caption{Chinese prompt and response templates of Claim-Query Generator.}
\label{tab:zh_prompt_cqg}
\end{table*}

\begin{table*}
\centering
\small
\begin{tabularx}{\textwidth}{X}
\toprule
English Prompt Template of Verifier\\
\midrule
\#\#\# Task\\
Given a claim and related reference searched by a query as input, your task is to determine whether the claim is valid based on the reference.\\
\\
\#\#\# Evaluation Criteria\\
Please make your judgment based on the following criteria and choose one of the three options:\\
1. Correct: The reference supports the claim.\\
2. Hallucination: The reference is relevant to the claim, but does not support the claim.\\
3. Irrelevant: The reference is irrelevant to the claim, thus does not contain enough information to determine the factuality of the claim. Only use this option when absolutely necessary.\\
\\
Provide only one option as the output. No additional explanation is allowed.\\
\\
\#\#\# Input\\
{[Claim]}\\
\{claim\}\\
{[Reference]}\\
\{reference\}\\
\bottomrule
\end{tabularx}
\caption{English prompt template of Verifier.}
\label{tab:en_prompt_verifier}
\end{table*}

\begin{table*}
\centering
\small
\begin{tabularx}{\textwidth}{X}
\toprule
Chinese Prompt Template of Verifier\\
\midrule
\#\#\# 任务\\
给定一条陈述以及由查询检索到相关的参考资料作为输入，你的任务是根据参考资料判断陈述是否成立。\\
\\
\#\#\# 判断标准\\
请依据以下标准进行判断，输出三个选项之一：\\
1. 正确：参考资料能够支持陈述。\\
2. 幻觉：参考资料与陈述相关，但并不支持陈述。\\
3. 无关：参考资料与陈述内容无关，信息不足，无法判断陈述的真实性。非必要不使用此选项。\\
\\
只输出一个选项，不允许有任何额外解释。\\
\\
\#\#\# 输入\\
{[陈述]}\\
\{claim\}\\
{[参考资料]}\\
\{reference\}\\
\bottomrule
\end{tabularx}
\caption{Chinese prompt template of Verifier.}
\label{tab:zh_prompt_verifer}
\end{table*}

\begin{table*}
\centering
\small
\begin{tabularx}{\textwidth}{X}
\toprule
English Prompt Template of LLM Baselines\\
\midrule
I want you act as an answer judge. Given a question and an answer, your objective is to determine if the provided answer contains non-factual or hallucinated information.\\
The response you give should be ``Yes" or ``No".\\
\\
\#\#\# Question\\
\{question\}\\
\#\#\# Answer\\
\{answer\}\\
\bottomrule
\end{tabularx}
\caption{English prompt template of LLM baselines.}
\label{tab:en_prompt_baseline}
\end{table*}

\begin{table*}
\centering
\small
\begin{tabularx}{\textwidth}{X}
\toprule
Chinese Prompt Template of LLM Baselines\\
\midrule
我希望你充当一个答案评审员。给定一个问题和一个答案，你的任务是判断所提供的答案中是否包含虚假或幻觉的信息。\\
你的回复应该为“是”或“否”。\\
\\
\#\#\# 问题\\
\{question\}\\
\#\#\# 答案\\
\{answer\}\\
\bottomrule
\end{tabularx}
\caption{Chinese prompt template of LLM baselines.}
\label{tab:zh_prompt_baseline}
\end{table*}

\end{CJK}

\end{document}